%% file: main.tex
\documentclass[10pt,a4paper]{article}

\usepackage[a4paper,top=22mm,bottom=23mm,left=25mm,right=25mm,headheight=20pt,headsep=7mm,footskip=11mm]{geometry}
\usepackage[T1]{fontenc}
\usepackage{XCharter}
\usepackage[charter]{newtxmath}
\usepackage[scale=0.90]{sourcesanspro}
\usepackage{microtype}
\usepackage{setspace}
\usepackage{graphicx}
\graphicspath{{figures/}}
\usepackage{caption}
\usepackage{subcaption}
\usepackage{booktabs}
\usepackage{array}
\usepackage{tabularx}
\usepackage[most]{tcolorbox}
\usepackage{amsmath}
\usepackage{mathtools}
\usepackage{algorithmic}
\usepackage{enumitem}
\usepackage{multirow}
\usepackage{longtable}
\usepackage{adjustbox}
\usepackage{svg}
\usepackage{tikz}
\usepackage{pgfplots}
\usepackage{fontawesome5}

\usepackage[table,dvipsnames]{xcolor}
\definecolor{reportcyan}{HTML}{C36A4A}
\definecolor{reportnavy}{HTML}{214E45}
\definecolor{reportink}{HTML}{24231F}
\definecolor{reportgray}{HTML}{746F66}
\definecolor{reportpanel}{HTML}{F4EFE5}
\definecolor{reportrule}{HTML}{D8CEBC}
\pgfplotsset{compat=1.18}
\usepackage{titlesec}
\usepackage{tocloft}
\usepackage{fancyhdr}
\usepackage{lastpage}
\usepackage[numbers,sort&compress]{natbib}
\usepackage{placeins}
\usepackage{hyperref}
\usepackage{bookmark}
\hypersetup{
  colorlinks=true,
  linkcolor=reportnavy,
  citecolor=reportnavy,
  urlcolor=reportcyan,
  pdftitle={SciHorizon-eLab: An Agentic Protocol-to-Task Compiler for Scalable Benchmarking of Scientific Embodied Agents},
  pdfauthor={Maokai Qin et al.},
  pdfkeywords={Scientific embodied agents, protocol-to-task compilation, benchmarking, simulation},
  bookmarksnumbered=true
}

\titleformat{\section}
  {\Large\bfseries\color{reportnavy}}
  {\fontsize{12}{12}\selectfont\sffamily\bfseries\color{reportcyan}\MakeUppercase{\thesection}}{1em}{}
\titlespacing*{\section}{0pt}{2.15em plus .3em minus .2em}{0.72em}
\titleformat{\subsection}
  {\large\bfseries\color{reportink}}
  {\sffamily\bfseries\color{reportcyan}\thesubsection}{0.75em}{}
\titlespacing*{\subsection}{0pt}{1.55em plus .2em minus .15em}{0.45em}
\titleformat{\subsubsection}
  {\normalsize\bfseries\color{reportink}}
  {\thesubsubsection}{0.65em}{}
\titlespacing*{\subsubsection}{0pt}{1.2em}{0.3em}

\setlist{topsep=0.45em,itemsep=0.2em,parsep=0pt,leftmargin=1.8em}
\setlist[itemize,1]{label=\textcolor{reportcyan}{\raisebox{0.15ex}{\scriptsize\ensuremath{\blacksquare}}}}
\renewcommand{\arraystretch}{1.18}
\newcolumntype{P}[1]{>{\raggedright\arraybackslash}p{#1}}

\providecommand{\fullmark}{%
  \tikz[baseline=-0.60ex]{\fill (0,0) circle (0.60ex);}% 
}
\providecommand{\partmark}{%
  \tikz[baseline=-0.60ex]{%
    \begin{scope}
      \clip (0,0) circle (0.60ex);
      \fill (-0.60ex,-0.60ex) rectangle (0,0.60ex);
    \end{scope}
    \draw[line width=0.50pt] (0,0) circle (0.60ex);
  }%
}
\providecommand{\emptymark}{%
  \tikz[baseline=-0.60ex]{\draw[line width=0.50pt] (0,0) circle (0.60ex);}% 
}

\newcommand{\Description}[1]{}

\renewcommand{\headrulewidth}{0.7pt}
\renewcommand{\footrulewidth}{0pt}
\renewcommand{\headrule}{\hbox to\headwidth{\color{reportrule}\leaders\hrule height \headrulewidth\hfill}}
\renewcommand{\footrule}{\hbox to\headwidth{\color{reportrule}\leaders\hrule height \footrulewidth\hfill}}
\fancypagestyle{plain}{%
  \fancyhf{}
  \fancyhead[L]{\scriptsize\sffamily\bfseries\color{reportnavy}\MakeUppercase{SciHorizon-eLab}}
  \fancyhead[R]{\scriptsize\sffamily\color{reportgray}\nouppercase{\leftmark}}
  \fancyfoot[L]{\scriptsize\sffamily\color{reportgray}Technical Report}
  \fancyfoot[R]{\scriptsize\sffamily\bfseries\color{reportcyan}\thepage\hspace{0.35em}\color{reportgray}/\hspace{0.35em}\pageref*{LastPage}}
  \renewcommand{\headrulewidth}{0.7pt}
  \renewcommand{\footrulewidth}{0pt}
}

\newif\ifshowtoc
\showtoctrue

\newcommand{\ModelName}{\textsc{SciHorizon-eLab}}
\newcommand{\BenchName}{SciVLABench}
\newcommand{\HIL}{human-agent coordination}
\newcommand{\reportlogo}{\includegraphics[height=0.48in]{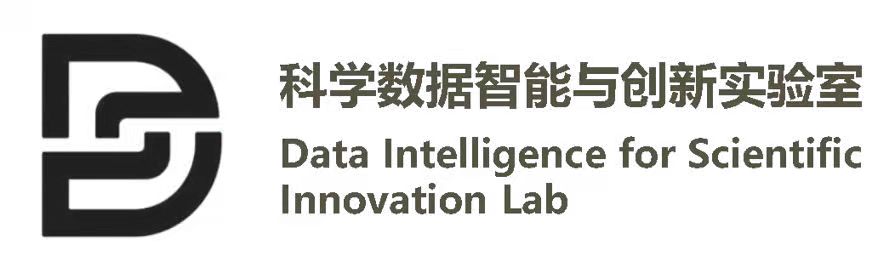}}
\newcommand{\reporttitle}{\ModelName: An Agentic Protocol-to-Task Compiler for Scalable Benchmarking of Scientific Embodied Agents}
\newcommand{\reportauthors}{Maokai Qin$^{1,2}$, Chuan Qin$^{1,\dagger, ^*}$, Qi Zhang$^{1, \dagger}$, Dianyu Liu$^{1}$, Zirui Liu$^{1}$, Hongting Niu$^{2}$, Yuanchun Zhou$^{1}$, Hengshu Zhu$^{1,^*}$}
\newcommand{\reportkeywords}{Scientific embodied agents, protocol-to-task compilation, benchmarking, simulation}

\newcommand{\custommaketitle}{%
  \thispagestyle{empty}
  \noindent\begin{minipage}[c]{0.66\textwidth}
    \reportlogo
  \end{minipage}%
  \begin{minipage}[c]{0.25\textwidth}
    \raggedleft\small\color{reportgray}\today
  \end{minipage}\par
  \vspace{0.8em}
  \noindent{\color{reportrule}\rule{\textwidth}{0.8pt}}\par
  \vspace{2.3em}
  \noindent{\sffamily\small\bfseries\color{reportcyan}\MakeUppercase{Technical Report}}\par
  \vspace{0.8em}
  \begin{flushleft}
    \sloppy
    {\fontsize{25}{29}\selectfont\bfseries\color{reportnavy}\reporttitle\par}
    \vspace{1.35em}
    {\large\bfseries\color{reportink}\reportauthors\par}
    \vspace{0.5em}
    
    {\small\sffamily\color{reportgray}$^{1}$Data Intelligence for Scientific Innovation Lab, Computer Network Information Center, Chinese Academy of Sciences\par}
    {\small\sffamily\color{reportgray}$^{2}$Beihang University\par}
    {\small\sffamily\color{reportgray}$^\dagger$Project Co-Lead \quad $^*$Corresponding Authors\par }
    \vspace{0.32em}
    {\footnotesize\sffamily\color{reportgray}\url{qinmaokai@buaa.edu.cn}; \url{chuanqin0426@gmail.com}; \url{zhangqi.fqz@gmail.com}; \url{liudianyu00@gmail.com}\par}
    {\footnotesize\sffamily\color{reportgray}\url{202311998114@mail.bnu.edu.cn}; \url{niuhongting@buaa.edu.cn}; \url{zyc@cnic.cn}; \url{zhuhengshu@gmail.com}\par}
  \end{flushleft}
  \vspace{1.15em}
}

\renewenvironment{abstract}{%
  \begin{tcolorbox}[
    enhanced,breakable,colback=reportpanel,colframe=reportpanel,
    boxrule=0pt,arc=1pt,left=1.15em,right=1.15em,top=0.95em,bottom=0.95em,
    borderline north={2pt}{0pt}{reportcyan}]
  \noindent{\sffamily\bfseries\color{reportcyan}\MakeUppercase{Abstract}}\hspace{0.65em}%
  \color{reportink}\ignorespaces
}{%
  \end{tcolorbox}
  \vspace{0.5em}
}

\begin{document}
\color{reportink}
\custommaketitle

\begin{abstract}
Embodied agents offer a promising route to automating scientific experimentation, yet their progress is constrained by the lack of reliable and systematic evaluation environments. Existing simulation-based laboratory benchmarks rely heavily on manual task engineering, making it challenging to systematically compile diverse scientific protocols into executable and verifiable embodied tasks at scale. To address this challenge, we introduce \mbox{\textbf{\ModelName}}, an agentic protocol-to-task compiler that formulates scientific embodied task construction as a compilation problem. Given a natural-language protocol of scientific experiments, \ModelName\ progressively compiles laboratory protocols into semantic-preserving embodied tasks through semantic grounding, executable task synthesis, and multi-stage simulation-based certification. The system generates semantically grounded environments, executable manipulation programs, and step-level success specifications, while enabling reproducible generation of expert demonstrations and execution traces. Using this pipeline, we further construct \BenchName, a ready-to-use benchmark comprising 300 certified tasks across diverse laboratory operations. It supports \HIL\ task execution, reproducible expert-demonstration generation, and ordered step-level evaluation. Across representative tasks, the strongest policy attains an average success rate of only 49.7\%, with further evaluations revealing pronounced weaknesses in human and embodied agent coordination. We publicly release the code, benchmark data, and evaluation toolkit at {\url{https://github.com/SciHorizon-elab/SciHorizon-elab}}.
\end{abstract}

{\noindent\small\sffamily\textbf{Keywords:} \reportkeywords\par}
\vspace{0.4em}

\ifshowtoc
  \clearpage
  \thispagestyle{fancy}
  \setcounter{tocdepth}{2}
  \tableofcontents
  \clearpage
\fi

\input{sections/01-introduction}

\input{sections/02-dataset}
\input{sections/03-methodology}

\input{sections/04-methods}
\input{sections/05-evaluation}

\FloatBarrier
\input{sections/06-conclusion}

\clearpage
\FloatBarrier
\input{sections/appendix}

\clearpage
\phantomsection
\pdfbookmark[1]{References}{references}
\addcontentsline{toc}{section}{References}
\bibliography{bibliography/reference}
\bibliographystyle{ACM-Reference-Format}

\end{document}

%% file: sections/01-introduction.tex
\section{Introduction}
\label{sec:introduction}

% Recent advances in multimodal perception, reasoning, and robot learning
% have accelerated the development of embodied agents capable of
% interacting with complex physical environments. Scientific laboratories
% represent a particularly demanding setting for such agents. Beyond
% visuomotor manipulation, laboratory procedures require agents to
% interpret ordered experimental instructions, identify the functional
% roles of instruments and materials, track intermediate states, satisfy
% safety constraints, and coordinate actions over extended procedural
% horizons. As embodied agents are increasingly envisioned as autonomous
% scientific assistants, reliable and reproducible evaluation of these
% capabilities has become an important research problem.

Recent advances in multimodal reasoning, foundation models, and robot learning have enabled embodied agents to interact with increasingly complex physical environments~\cite{brohan2023rt1,zitkovich2023rt2,driess2023palme}. Scientific laboratories are a particularly demanding domain for such agents: successful experimentation requires not only visuomotor manipulation, but also the interpretation of procedural instructions, an understanding of the functional roles of instruments and materials, the tracking of intermediate experimental states, and adherence to scientific and safety constraints over long action horizons~\cite{king2009automation,szymanski2023alab}. As embodied agents emerge as promising assistants for scientific experimentation, their progress increasingly depends on reliable and systematic evaluation environments.

Physics-based simulation provides a safe, controllable, and scalable setting for evaluating scientific embodied agents~\cite{todorov2012mujoco,savva2019habitat,shen2021igibson}, but constructing simulation-ready laboratory tasks remains difficult. Scientific protocols are written for trained human experimenters and typically describe procedures at a high level, whereas embodied benchmarks require explicit representations of entities, actions, states, constraints, and measurable success conditions. Transforming a protocol involving operations such as transferring, heating, or mixing into an executable benchmark task therefore requires more than translating its instructions. The protocol's scientific intent must be grounded in a simulation environment, realized as an executable manipulation procedure, and expressed through objective, process-level evaluation specifications. Existing embodied and laboratory benchmarks commonly rely on manually authored or user-configured task structures~\cite{li2025chemistry3d,li2025labutopia,liu2026pipette}, while recent generative approaches automate individual components such as scene creation, task proposal, or controller synthesis~\cite{deitke2022procthor,chen2025robotwin2,ren2026labvla}. However, these components alone do not provide an end-to-end, protocol-grounded construction process that jointly produces executable tasks and independent evaluation specifications, and then certifies their consistency through simulation before benchmark inclusion.

To bridge this gap, we introduce \textbf{\ModelName}, an agentic protocol-to-task compiler that transforms natural-language scientific protocols into executable and verifiable embodied tasks. Its compilation pipeline comprises three stages: (1)~\textsl{Semantic-Preserving Environment Compilation}, which grounds a protocol in a simulator-compatible environment and a structured task representation; (2)~\textsl{Executable Task Specification Generation}, which synthesizes an executable manipulation program and an independently defined step-level success specification; and (3)~\textsl{Simulation Certification and Task Instantiation}, which validates structural correctness, physical executability, and ordered success conditions through simulation before instantiating certified tasks. By unifying semantic grounding, executable synthesis, and simulation-based certification, \ModelName\ enables scalable and reproducible construction of protocol-grounded evaluation tasks.

Using \ModelName, we construct \textbf{\BenchName}, a pre-generated and ready-to-use benchmark for scientific embodied agents. \BenchName\ organizes 51 protocol-grounded source scenarios into five primary operation families. Using a library of 30 registered atomic manipulation skills, these scenarios are realized as 300 simulation-certified tasks. Each task contains a grounded environment, an executable reference program, and an ordered step-level success specification, enabling reproducible, on-demand generation of expert demonstrations and execution traces. Certified randomization of poses, layouts, initial states, and visual conditions further yields diverse task instances without changing the task's procedural or evaluation semantics. In addition to autonomous execution tasks, \BenchName\ includes 15 certified human-agent collaborative tasks. It supports binary task-success assessment, ordered step-level evaluation, and same-task randomized evaluation.

We conduct extensive experiments to evaluate both the benchmark and the underlying compiler. On ten representative certified tasks, we compare three visuomotor policies using task-level success and ordered step-level completion; the strongest policy achieves an average task success rate of only 49.7\%, highlighting the difficulty of scientific manipulation. We further assess \HIL\ tasks and transfer to geometrically distinct object bindings, exposing substantial limitations in current policies. Finally, a controlled audit of 100 task commands measures the first-pass reliability, recoverability, and remaining manual effort of \ModelName. To support reproducible research and rapid iteration, we publicly release the code, benchmark data, and evaluation toolkit at {\url{https://github.com/SciHorizon-elab/SciHorizon-elab}}. Our contributions are summarized as follows:
\begin{itemize}[leftmargin=*]
\item We formulate scientific embodied task construction as a
protocol-to-task compilation problem, providing a systematic path from
natural-language protocols to executable and objectively verifiable tasks.
\item We introduce \textbf{\ModelName}, an agentic compiler that integrates
semantic-preserving environment compilation, independently specified
execution and evaluation logic, and multi-stage simulation-based certification.
\item We construct \textbf{\BenchName}, a protocol-grounded benchmark of
300 certified tasks, together with reproducible expert-demonstration
generation and evaluation protocols for procedural completion, transfer, and
{\HIL}.
\end{itemize}

\begin{center}
    \vspace{1mm}
    \captionof{table}{Comparison of representative embodied-manipulation benchmarks and automated task-construction systems.}
    \label{tab:benchmark-comparison}
    \scriptsize
    \setlength{\tabcolsep}{0pt}
    \renewcommand{\arraystretch}{1.12}
    \newcommand{\tabfull}{\raisebox{0.18ex}{\scalebox{1.15}{$\bullet$}}}
    \newcommand{\tabempty}{\raisebox{0.12ex}{\scalebox{1.25}{$\circ$}}}
    \newcommand{\tabpart}{\raisebox{0.15ex}{\scalebox{0.8}{$\odot$}}}
    \begin{tabular*}{\textwidth}{@{\extracolsep{\fill}}lcccccccccc}
        \toprule
        \multirow{2}{*}{System}
        & \multirow{2}{*}{Role}
        & \multirow{2}{*}{Sci. Lab.}
        & \multicolumn{3}{c}{Benchmark Evaluation}
        & \multicolumn{5}{c}{Task Construction} \\
        \cmidrule(lr){4-6}
        \cmidrule(lr){7-11}
        & & &
        {\footnotesize Ord. Step}
        & {\footnotesize Bind. Transfer}
        & {\footnotesize H--A Coord.}
        & {\footnotesize Protocol}
        & {\footnotesize Auto Task}
        & {\footnotesize Indep. Eval.}
        & {\footnotesize Sim. Cert.}
        & {\footnotesize Auto Demo} \\
        \midrule
        RLBench~\cite{james2020rlbench}
        & B & \tabempty & \tabempty & \tabpart & \tabempty
        & \tabempty & \tabempty & \tabempty & \tabpart & \tabfull \\
        LIBERO~\cite{liu2023libero}
        & B+G & \tabempty & \tabempty & \tabpart & \tabempty
        & \tabempty & \tabfull & \tabempty & \tabempty & \tabempty \\
        VLABench~\cite{zhang2024vlabench}
        & B & \tabempty & \tabpart & \tabfull & \tabempty
        & \tabempty & \tabempty & \tabempty & \tabempty & \tabfull \\
        LabUtopia~\cite{li2025labutopia}
        & B & \tabfull & \tabpart & \tabfull & \tabempty
        & \tabempty & \tabempty & \tabempty & \tabempty & \tabfull \\
        Pipette~\cite{liu2026pipette}
        & B+G & \tabfull & \tabempty & \tabempty & \tabempty
        & \tabempty & \tabpart & \tabempty & \tabempty & \tabpart \\
        RoboGen~\cite{wang2024robogen}
        & G & \tabempty & \tabempty & \tabempty & \tabempty
        & \tabempty & \tabfull & \tabempty & \tabpart & \tabfull \\
        RoboTwin 2.0~\cite{chen2025robotwin2}
        & B+G & \tabempty & \tabempty & \tabempty & \tabempty
        & \tabempty & \tabfull & \tabempty & \tabpart & \tabfull \\
        RoboGenesis (LabVLA)~\cite{ren2026labvla}
        & G & \tabfull & \tabpart & \tabpart & \tabempty
        & \tabpart & \tabfull & \tabempty & \tabpart & \tabfull \\
        \midrule
        \rowcolor{gray!8}
        \textbf{\ModelName}
        & \textbf{B+G} & \tabfull & \tabfull & \tabfull & \tabfull
        & \tabfull & \tabfull & \tabfull & \tabfull & \tabfull \\
        \bottomrule
    \end{tabular*}

    \vspace{3pt}

    \begin{minipage}{\textwidth}
        \scriptsize
        \textit{Notes.}
        B/G denote benchmark/generator; \fullmark/\partmark/\emptymark denote reported full, partial or related, and unreported support.
        Scientific-Laboratory (Sci. Lab.) denotes scientific-laboratory specialization.
        Ordered-Step Evaluation (Ord. Step) denotes ordered process-level evaluation.
        Binding Transfer (Bind. Transfer) denotes evaluation with unseen object or instrument bindings under unchanged task semantics.
        Human--Agent Coordination (H--A Coord.) denotes coordination with hidden-timing external interventions.
        Protocol Input (Protocol) denotes traceable scientific-protocol input.
        Automatic Task Construction (Auto Task) denotes automatic construction of executable task definitions.
        Independent Evaluation (Indep. Eval.) denotes success criteria generated separately from execution.
        Simulation Certification (Sim. Cert.) denotes simulation validation before release.
        Automatic Demonstration Generation (Auto Demo) denotes automatic generation of expert demonstrations.
    \end{minipage}
    \vspace{-3mm}
\end{center}

%% file: sections/02-dataset.tex
\section{Related Work}
\label{sec:related-work}

\subsection{Scientific Embodied Agents}
\label{sec:related-lab}

Automated laboratory systems are becoming an important pathway toward more reproducible and autonomous scientific experimentation. Representative systems such as Chemputer~\cite{steiner2019organic}, Artificial Chemist~\cite{epps2020artificial}, Synbot~\cite{ha2023aidriven}, and MARS-Chem~\cite{dai2024autonomous} integrate robotic hardware with structured experimental workflows for chemical synthesis, materials exploration, and reaction optimization. These systems demonstrate the feasibility of machine-executed science, but are generally designed around specialized hardware, preconfigured procedures, and domain-specific objectives.

To reduce the cost and safety risks of physical experimentation, recent work has also explored scientific embodied agents in simulation. Chemistry3D~\cite{li2025chemistry3d} supports robotic manipulation together with observable scientific states such as temperature, color, pH, transparent objects, and liquid interactions. LabUtopia~\cite{li2025labutopia} provides interactive laboratory assets, procedural scene generation, and hierarchical evaluation from atomic manipulation to long-horizon experiments. Pipette~\cite{liu2026pipette} further offers editable wet-lab assets, extensible task registration, and simulation-based data augmentation. These platforms provide valuable environments for scientific embodied learning, but their tasks are primarily manually specified or user-configured rather than systematically compiled from natural-language scientific protocols.

\subsection{Embodied Agent Benchmarking}
\label{sec:related-benchmarks}

Reliable benchmarking is essential for measuring the capabilities and failure modes of embodied agents. Real-robot systems such as RT-1~\cite{brohan2023rt1} and RT-2~\cite{zitkovich2023rt2} evaluate visuomotor policies in physical manipulation settings, but such evaluation is costly, difficult to standardize, and constrained by hardware and safety requirements. Simulation therefore provides a more controllable and reproducible alternative. RLBench~\cite{james2020rlbench} and LIBERO~\cite{liu2023libero} provide standardized manipulation suites with expert demonstrations, ManiSkill2~\cite{gu2023maniskill2} studies generalization across object geometries and physical interactions, and CALVIN~\cite{mees2022calvin} evaluates long-horizon skill composition. VLABench~\cite{zhang2024vlabench} extends evaluation to visual, spatial, physical, and long-horizon reasoning, while LabUtopia~\cite{li2025labutopia} focuses specifically on scientific laboratory environments.

To reduce the cost of manually constructing tasks and demonstrations, RoboGen~\cite{wang2024robogen}, RoboTwin 2.0~\cite{chen2025robotwin2}, and RoboGenesis~\cite{ren2026labvla} automate components such as task proposal, scene construction, execution-program synthesis, and rollout collection. However, as summarized in Table~\ref{tab:benchmark-comparison}, their task definitions and evaluation logic are not systematically derived from source scientific protocols, and successful execution alone does not guarantee semantic preservation, independently specified process-level criteria, or consistent certification before benchmark inclusion. \ModelName\ addresses this gap by compiling scientific protocols into grounded environments, executable procedures, and independent step-level success specifications, followed by simulation-based certification. Based on this framework, \BenchName\ provides 300 certified scientific embodied tasks with reproducible expert demonstrations, ordered procedural evaluation, held-out object-binding evaluation, and human-agent coordination tasks.

\begin{center}
  \vspace{2mm}
  \includegraphics[width=\textwidth]{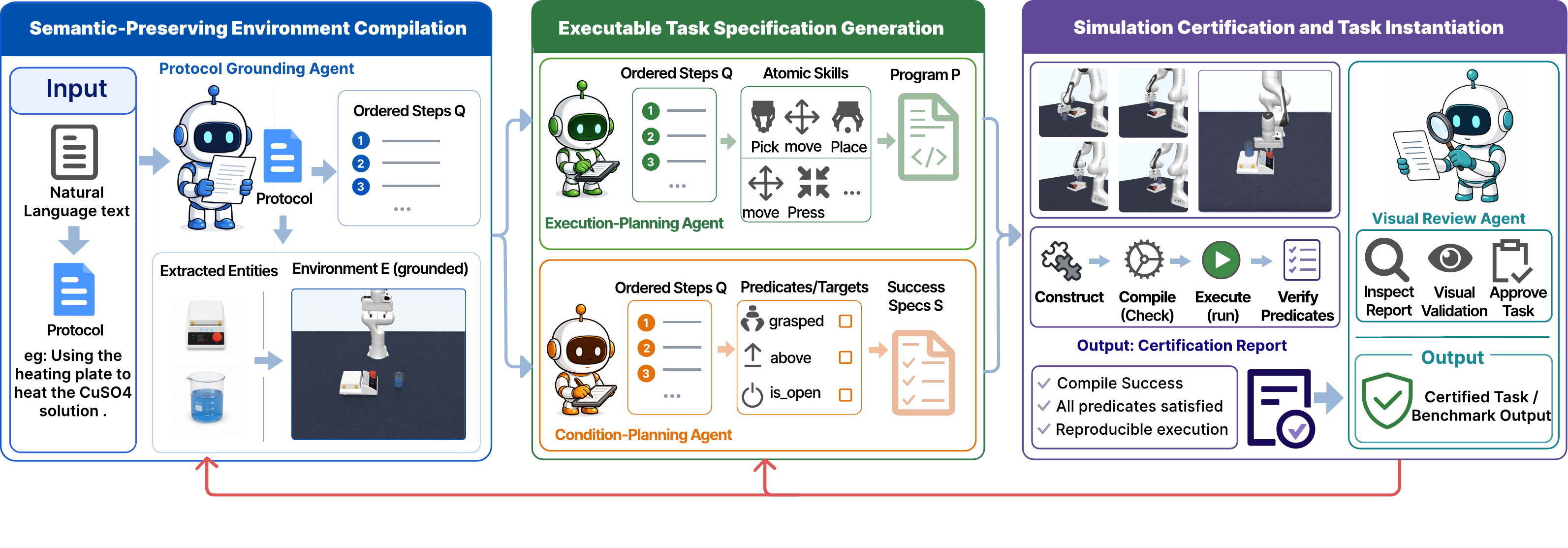}
  \captionof{figure}{Overview of the \ModelName\ protocol-to-task compilation framework. Given a natural-language scientific protocol, \ModelName\ progressively compiles it into a simulator-ready embodied task through semantic-preserving environment compilation, executable task specification generation, and simulation-based certification and task instantiation.}
  \Description{An experimental protocol passes through three compilation stages to produce grounded environments, executable task specifications, simulation-certified \textcolor{blue}{templates}, and randomized benchmark instances.}
  \label{fig:method-overview}
\end{center}

\FloatBarrier

%% file: sections/03-methodology.tex
\section{\ModelName}
\label{sec:method}

A scientific protocol $p$ describes a laboratory procedure in natural language, including its experimental entities, ordered operations, and observable outcomes.
Our goal is to transform such a protocol into an embodied task that can be instantiated, executed, and automatically evaluated in a physics-based environment.

We formulate this protocol-to-task compilation problem as
\begin{equation}
\mathsf{C}(p)=\tau,
\end{equation}
where $\mathsf{C}$ denotes the proposed compiler.
To support both execution and evaluation, the resulting task $\tau$ contains three components:
\begin{equation}
\tau = (E, P, S),
\end{equation}
where $E$ represents the grounded simulation environment, $P$ specifies the robot-executable procedure, and $S$ specifies the ordered conditions used to evaluate task completion.
The central challenge is to derive these mutually consistent components from $p$ while preserving its explicit scientific semantics and completing the robot-specific details that are omitted from human-oriented protocols.

\subsection{Framework Overview}
\label{sec:method-overview}

% To transform a natural-language experimental protocol $p$ into an executable and verifiable embodied task, we propose SciHorizon-Lab, an automated protocol-to-task compilation framework. Rather than generating simulator code directly from the protocol, the framework constructs typed intermediate representations that make protocol semantics, action realization, and task evaluation explicit.

% The compilation is formalized as
% \begin{equation}
% \mathcal{C}(p)
% =
% \tau
% =
% (\mathcal{E},\mathcal{P},\mathcal{S}),
% \end{equation}
% where $\mathcal{E}$ is a grounded environment specification, $\mathcal{P}$ is an executable skill program, and $\mathcal{S}$ is an independently generated step-level success specification.
% Shared entity and step identifiers maintain alignment among the three components,whereas the separation between $\mathcal{P}$ and $\mathcal{S}$ decouples action generation from success judgment. Only templates that can be compiled, executed, and verified in simulation are retained for benchmark instantiation.
As illustrated in Figure~\ref{fig:method-overview}, \ModelName\ transforms a natural-language scientific protocol into a simulator-ready embodied task through three stages.
First, \textbf{Semantic-Preserving Environment Compilation} interprets the protocol and constructs the simulation environment required to realize its experimental procedure.
Second, \textbf{Executable Task Specification Generation} completes the robot-specific operational details and independently derives the criteria used to evaluate task completion.
Third, \textbf{Simulation Certification and Task Instantiation} compiles and executes the generated task in simulation, verifies its procedural outcomes, and instantiates qualified tasks under admissible randomization.
The following subsections describe these three stages in detail.

\FloatBarrier
\subsection{Semantic-Preserving Environment Compilation}
\label{sec:environment-compilation}

% 把自然语言实验协议中与环境有关的语义，转换为一个可以被模拟器实例化、被后续规划模块引用、并满足基本物理可行性的结构化环境。

%协议解析→实体规范化→资产与状态实例化→场景构建与验证

% 协议中有哪些实体、属性、操作步骤和约束？
% 不同表达如何规范化为全局一致的实体记录？
% 哪些实体需要独立的三维资产，哪些应表示为容器中的内容物或状态？
% 这些实体如何被实例化、布置并验证为一个可执行场景？

The first stage converts the source protocol $p$ into an ordered sequence of grounded protocol steps $Q$ and a simulator-compatible environment $E$.
We denote the output of this stage by
\begin{equation}
\mathsf{Ground}(p)=(Q,E).
\end{equation}
The sequence $Q=(q_1,\ldots,q_m)$ preserves the procedural order and explicit experimental operations described in $p$, while $E$ records the entities, asset bindings, relations, and initial-state constraints required to realize these operations in simulation.
%In this stage, semantic preservation means that each task-relevant element in $Q$ and $E$ remains traceable to an explicit description in the source protocol.
%The compiler preserves the stated entities, attributes, action verbs, modifiers, spatial relations, and procedural constraints, while postponing robot-specific action completion to the next stage.
%This separation prevents low-level manipulation decisions from being introduced as if they were part of the original scientific procedure.
In this stage, semantic preservation requires that every task-relevant element in $Q$ and $E$ remains explicitly traceable to the source protocol. The compiler retains the original entities, attributes, action descriptions, modifiers, spatial relations, and procedural constraints, while deferring robot-specific execution details to the subsequent stage. This separation ensures that low-level manipulation decisions are not conflated with the scientific procedure specified by the protocol.

A schema-constrained \textbf{grounding agent} first identifies the entities and ordered operations explicitly described in the protocol.
Each grounded step $q_i$ preserves its source operation, referenced entities, and stated constraints.
% References to the same experimental entity are assigned a shared task-level identifier across all steps, preventing changes in object identity or role as the protocol is processed.
% The structured output is accepted only when every entity reference can be resolved and the original procedural order is preserved.
Mentions of the same experimental entity are assigned a consistent task-level identifier across all steps, preventing unintended changes in object identity or functional role during protocol grounding. The structured representation is accepted only if all entity references are successfully resolved and the original procedural order is strictly maintained.

% The grounded entities are then mapped to simulator representations according to their roles in the protocol.
% Instruments, containers, and other manipulable objects are bound to compatible simulator assets.
% Liquids, powders, and similar materials are represented as contents or states associated with their physical carriers rather than as independent manipulable objects.
% Asset retrieval and binding are constrained by the semantic category and interaction requirements of each entity.
% The compiler may add simulator-specific information, such as collision geometry, interaction sites, or initialization parameters, but these additions are used only to support simulation and do not modify the protocol-level entities, operations, or intended outcomes.

The grounded entities are subsequently instantiated into simulator representations according to their functional roles defined in the protocol. Instruments, containers, and other manipulable objects are mapped to compatible simulator assets that support the required interactions. Materials such as liquids and powders are modeled as contents or internal states associated with their physical carriers rather than as independent manipulable entities. Asset retrieval and binding are constrained by each entity’s semantic category and interaction requirements. The compiler may augment these representations with simulator-specific properties, including collision geometry, interaction affordances, and initialization parameters, but such additions serve only to enable executable simulation and do not alter the protocol-level entities, operations, or intended experimental outcomes.

Finally, the grounded relations and initialization constraints are used to assemble the initial scene.
Deterministic checks ensure that all references remain consistent, selected assets support the required interactions, and initial spatial and state relations agree with the protocol.
The resulting pair $(Q,E)$ therefore provides a traceable simulation representation of the source procedure.
The next stage uses this representation to determine how each grounded step $q_i$ should be executed and how its completion should be evaluated.

\subsection{Executable Task Specification Generation}
\label{sec:task-specification}

% 把3.2得到的结构化输入：(Q,E)
% 转化为一个可执行、可验证的任务规范：
% (P,S).
% 其中：
% P回答：机器人应该怎样完成协议步骤？
% S回答：怎样判断每个步骤是否成功？

The second stage converts the grounded procedure and environment produced by the previous stage into the two task components required for execution and evaluation:
\begin{equation}
\mathsf{Specify}(Q,E)=(\mathcal{P},\mathcal{S}).
\end{equation}
Let $\mathcal{P}=(P_1,\ldots,P_m)$ and $\mathcal{S}=(s_1,\ldots,s_m)$, where $\mathcal{P}$ denotes the executable skill program and $\mathcal{S}$ denotes the step-level success specification.
Each grounded step $q_i$ serves as the shared alignment unit: ${P}_i$ specifies how the step is executed, while $s_i$ defines the observable outcome required to verify its success.
This design ensures that execution and evaluation are consistently grounded in the same entities, operations, and procedural positions.

\subsubsection{Executable Procedure Generation.}
An \textbf{execution-planning agent} generates $\mathcal{P}$ using the complete grounded sequence $Q$, the entity and asset capabilities recorded in $E$, and a registered library of atomic laboratory skills.
The agent plans over the full procedure rather than treating each step independently.
This allows it to track intermediate object and device states across steps and to insert robot-specific operations that are implicit in the human-oriented protocol, such as approaching, grasping, releasing, or repositioning an object.
These additions operationalize the stated procedure without changing its experimental step order or intended outcomes.

Each $P_i$ is represented as an ordered sequence of parameterized atomic skills whose arguments refer to the shared task-level entity identifiers.
A deterministic validator checks that the generated program preserves the order of $Q$, invokes only registered skills, supplies all required parameters, and uses entity bindings supported by the grounded environment $E$.

\subsubsection{Success-Condition Generation.}
% In parallel, a separately prompted \textbf{condition-generation agent} derives $\mathcal{S}$ from $Q$, $E$, and a registered library of verification predicates.
% For each $q_i$, it translates the intended observable outcome $s_i$ into one or more conditions over simulator states or events.
% These conditions may describe spatial relations, object states, material or device states, and temporal requirements such as waiting for an external trigger or avoiding premature interaction.
In parallel, an independently prompted \textbf{condition-planning agent} derives $\mathcal{S}$ from $Q$, $E$, and a registered library of verification predicates. For each $q_i$, it converts the intended observable outcome into executable verification conditions over simulator states or events. The resulting specifications capture multiple dimensions of task success, including spatial relations, object states, material or device states, and temporal constraints such as delayed execution, external triggers, or prevention of premature interactions.

% The condition-generation agent does not receive the generated program $\mathcal{P}$.
% Its conditions are therefore derived from the grounded protocol rather than from the specific actions selected by the execution planner.
% A deterministic validator checks that every grounded step $q_i$ has a corresponding condition $s_i$, that all predicate types and arguments are supported, and that all entity references are consistent with $E$.
The condition-planning agent does not have access to the generated program $\mathcal{P}$. Therefore, the verification conditions are derived solely from the grounded protocol rather than from the actions produced by the execution planner. A deterministic validator verifies that each grounded step $q_i$ is associated with a corresponding condition $s_i$, that all predicate types and arguments are supported, and that all entity references remain consistent with~$E$.

% The two agents share the grounded procedure and environment but not their generated outputs.
% This design maintains step- and entity-level alignment between $\mathcal{P}$ and $\mathcal{S}$ while separating action generation from success judgment.
% The resulting pair $(\mathcal{P},\mathcal{S})$ is passed to the certification stage, where simulation execution and complementary verification procedures determine whether the complete task template is runnable and consistent with the source protocol.
The two agents share the same grounded procedure and environment but independently generate their respective outputs. This design preserves step- and entity-level alignment between $\mathcal{P}$ and $\mathcal{S}$ while decoupling action synthesis from success verification. The resulting pair $(\mathcal{P},\mathcal{S})$ is then passed to the certification stage, where simulation-based execution and complementary verification procedures assess whether the generated tasks is executable, reproducible, and consistent with the source protocol.

\subsection{Simulation Certification and Task Instantiation}
\label{sec:certification}

Given the task $\tau$ and its aligned source information $(p,Q)$, the final stage determines whether the generated specification forms a runnable and verifiable task in the MuJoCo physics engine~\cite{todorov2012mujoco}. The stage first compiles and executes the task, then evaluates the resulting execution through complementary state-based and visual-semantic evidence. Only tasks that pass all required checks are certified and used to generate benchmark instances.

\subsubsection{Task Compilation and Physical Execution.}
A deterministic compiler transforms $\tau$ into simulator-native task and configuration classes.
Shared entity identifiers are preserved when binding assets, skill arguments, and verification conditions to simulator objects.
Before registration, the generated classes undergo deterministic validation to verify syntax correctness and required class and method interfaces. Skill and condition parameters are produced by dedicated upstream planning stages and bound to simulator entities through preserved UID references.

After successful registration, the program $\mathcal{P}$ is then executed in \mbox{MuJoCo}, producing an execution trace $\xi=(x_0,\ldots,x_H)$,
% \begin{equation}
% \xi=(x_0,\ldots,x_H),
% \end{equation}
where $x_t$ records the simulator state, skill-execution status, robot waypoints, and multimodal observations at trace index $t$.
Registration or execution failures prevent the current candidate from entering task verification and are recorded as structured diagnostics for subsequent inspection or refinement.

\subsubsection{Complementary Task Verification.}
% A successful execution is evaluated through two complementary verification channels.
% First, the \textbf{predicate verifier} evaluates the conditions in $\mathcal{S}$ using simulator states recorded at the completion of each grounded step $q_i$.
% State-based conditions are checked at the corresponding step boundary, while temporal conditions are evaluated over the relevant execution segment.
% The dependency relation $\prec$ ensures that the conditions are satisfied in the order required by the source procedure.
A successful execution is verified through two complementary verification channels. First, the predicate verifier evaluates the specifications in $\mathcal{S}$ using simulator states recorded at the completion of each grounded step $q_i$. State-based predicates are checked at the corresponding step boundaries, whereas temporal predicates are evaluated over their associated execution segments. The dependency relation $\prec$ enforces that all predicates are satisfied in the procedural order required by the source protocol.
Second, a \textbf{visual-review agent} assesses step-aligned keyframes and multiview observations with respect to the source protocol and the expected outcomes of each step.
The keyframes include intermediate observations from the atomic skills associated with each $q_i$, as well as the final observation collected at the completion of $q_i$.

Predicate verification checks properties explicitly represented in the simulator, whereas visual review assesses observable procedural semantics that the predicate library cannot fully capture.
% Requiring both forms of evidence reduces the risk that a task passes certification while realizing an unintended procedure or outcome.
% Requiring both forms of evidence reduces the risk of certifying tasks that satisfy low-level checks but deviate from the intended procedure or outcome.
Requiring both forms of evidence ensures that certified tasks are not only executable but also faithful to the intended scientific procedure.

\subsubsection{Certification and Task Instantiation.}
Let $\delta_{\mathrm{comp}}$, $\delta_{\mathrm{exec}}$, $\delta_{\mathrm{pred}}$, and $\delta_{\mathrm{vis}}$ indicate successful compilation and registration, physical execution, predicate verification, and visual-semantic review, respectively.
We define the certification result as
\begin{equation}
\Gamma(\tau)
=
\mathbb{I}
\left[
\delta_{\mathrm{comp}}
\land
\delta_{\mathrm{exec}}
\land
\delta_{\mathrm{pred}}
\land
\delta_{\mathrm{vis}}
\right].
\end{equation}
A task is certified when $\Gamma(\tau)=1$.

Each certified task $\tau$ can then generate task instances by sampling from its admissible initialization and randomization ranges.
The sampled parameters may vary object poses, workspace layouts, initial states, and visual conditions, while preserving the protocol-level object roles, execution program, and success conditions.
The resulting certified tasks and randomized instances constitute the task resources described in Section~\ref{sec:benchmark}.

%% file: sections/04-methods.tex
\section{\BenchName}
\label{sec:benchmark}

Using \ModelName, we construct \BenchName, a ready-to-use benchmark for training and evaluating scientific embodied agents. The benchmark consists of 300 certified tasks for autonomous execution and 15 \HIL\ tasks.

Each certified task retains the grounded environment, executable expert program, and ordered success specifications generated by \ModelName. Researchers can therefore directly instantiate tasks, generate expert demonstrations, and evaluate embodied policies without rerunning the complete protocol-to-task compilation process. The remainder of this section describes the task sources and benchmark scope, followed by the released resources and supported evaluation protocols.

\subsection{Task Sources and Benchmark Scope}
\label{sec:benchmark-scope}

We curate laboratory procedures from scientific literature, automated laboratory systems, wet-lab robotic benchmarks, and laboratory safety procedures. We use a source scenario as the basic unit of task curation. Each source scenario describes a coherent operation-level procedure with explicitly identifiable entities, an ordered sequence of embodied actions, and observable intermediate or final outcomes. A source scenario is retained when its entities and ordered operations can be explicitly identified, its core interactions can be grounded to available or extensible simulator assets and manipulation skills, and its outcomes can be evaluated through simulator states or visual observations.
% For procedures whose scientific outcomes depend on microscopic chemical or biological mechanisms beyond the scope of the simulator, we preserve and evaluate the corresponding embodied operations rather than attempting to reproduce the underlying reaction mechanisms.
% Complete source provenance, scenario extraction rules, and inclusion and exclusion criteria are provided in Appendix~\ref{app:benchmark-sources}.

% The resulting source pool contains 51 protocol-grounded scenarios covering five major laboratory operation families: \emph{Liquid Handling and Transfer}, \emph{Mixing and Agitation}, \emph{Solid Handling and Weighing}, \emph{Thermal Control and Incubation}, and \emph{Apparatus and Workspace Interaction}. These families summarize the operational coverage of the current benchmark rather than define an exhaustive taxonomy of scientific experimentation.
% Detailed task-to-family assignments and family-wise statistics are provided in Appendix~\ref{app:benchmark-composition}.

The resulting source pool contains 51 protocol-grounded source scenarios into a taxonomy consisting of five primary operation families. The five families include \emph{Liquid Handling and Transfer}, \emph{Mixing and Agitation}, \emph{Solid Handling and Weighing}, \emph{Thermal Control and Incubation}, and \emph{Apparatus and Workspace Interaction}. The operation families provide high-level coverage of laboratory workflows. This taxonomy is designed to characterize the operational scope of the benchmark rather than serve as an exhaustive categorization of all scientific experimentation.

% Supported by a library of 30 registered atomic manipulation skills, the source scenarios are compiled into 300 simulation-certified embodied tasks. Each task can be instantiated under admissible variations of initial object poses, workspace configurations, and other task-defined conditions while preserving its underlying operation, executable procedure, and success specification.
% This design supports repeated evaluation under diverse initial configurations without regenerating or redefining the underlying task.
With a library of 30 registered atomic manipulation skills, the source scenarios are compiled into 300 simulation-certified embodied tasks. Each task can be instantiated under admissible variations of initial object poses, workspace configurations, and other task-specific conditions while preserving its underlying operation semantics, executable procedure, and success specification. This design enables repeated evaluation across diverse initial configurations without requiring task regeneration or manual redefinition.
% The definitions of the five operation families, their task distributions, and the complete task and skill manifests are provided in Appendix~\ref{app:benchmark-details}.

\noindent \textbf{Human-Agent Coordination Tasks.} 
% A dedicated subset of 15 certified tasks focuses on \HIL. In these tasks, progress depends on an externally triggered change in the experimental environment, such as the ignition of a laboratory lamp or the addition of material to a container. The agent must avoid premature interaction, recognize the relevant change from its observations, and resume the remaining procedure only after the required condition has been satisfied. This subset enables controlled evaluation of safe waiting, event recognition, post-intervention response, and final task completion in interactive scientific experimentation settings.
Existing embodied laboratory benchmarks primarily focus on autonomous task execution, overlooking the coordination challenges that arise when experimental progress depends on external human interventions or environmental events. To address this gap, our \BenchName\ introduces a dedicated subset of 15 certified tasks for scalable evaluation of human-agent coordination in scientific workflows. These tasks are automatically compiled from protocol-defined dependencies where execution cannot proceed until an externally triggered condition is satisfied, such as the ignition of a laboratory lamp or the addition of material to a container. The agent must recognize the relevant event from its observations, safely defer interaction before the condition is met, and resume the remaining procedure only after the required state transition occurs. This design enables controlled evaluation of coordination capabilities, including safe waiting, event understanding, intervention response, and reliable completion of interactive scientific procedures.

\FloatBarrier
\subsection{Released Resources and Evaluation Protocols}
\label{sec:released-resources}

% Each certified task is released as a structured package containing a natural-language instruction, a grounded environment specification, an executable expert program, ordered step-level success specifications, and certification metadata. Executing the expert program under admissible randomized initial conditions produces expert demonstrations, multimodal observations, execution traces, keyframes, and step-level verification records. Demonstrations can therefore be generated reproducibly on demand with different random seeds, rather than being restricted to a fixed collection of stored trajectories.
Each certified task is released as a structured package containing a natural-language instruction, a grounded environment specification, an executable expert program, ordered step-level success specifications, and certification metadata. By executing the expert program under admissible randomized initial conditions, the benchmark can reproducibly generate expert demonstrations, multimodal observations, execution traces, keyframes, and step-level verification records. This design enables on-demand demonstration generation with different random seeds, avoiding dependence on a fixed set of pre-recorded trajectories and supporting scalable evaluation under diverse task configurations.
% Detailed task schemas, metadata fields, and randomization parameters are provided in Appendix~\ref{app:benchmark-details}.

The ordered success specifications support evaluation at both the task and procedural levels. Complete task success requires all step-level conditions to be satisfied in the prescribed order, while step-wise evaluation measures partial procedural completion when an agent fails before finishing the full task. \HIL\ tasks additionally record premature interactions and post-trigger responses, allowing pre-intervention safety and event-conditioned execution to be assessed separately from final task completion. Formal metric definitions are provided in Section~\ref{sec:experiments}.

\BenchName\ supports three complementary evaluation protocols. \emph{Same-task randomized evaluation} trains and evaluates a policy on independently sampled instances of the same certified task, measuring robustness to variations in initial poses and workspace configurations. \emph{Held-out object-binding evaluation} replaces a training object with a geometrically distinct object while preserving the high-level operation and success specification, measuring transfer across laboratory object bindings. \emph{\HIL\ evaluation} randomizes the timing of an external intervention and requires the policy to infer the resulting state change from its observations before resuming execution.
% The task selections, training and evaluation splits, episode budgets, and implementation details for these protocols are described in Section~\ref{sec:experiments} and Appendix~\ref{app:benchmark-details}.

%% file: sections/05-evaluation.tex
\section{Experiments}
\label{sec:experiments}
We evaluate two complementary aspects of our work.
First, we assess the evaluation utility of \BenchName\ by comparing representative visuomotor policies on certified laboratory tasks and examining their procedural completion, human-agent coordination, and generalization to held-out object bindings.
Second, we assess the practical reliability of \ModelName\ through a controlled audit of its compilation and certification process.
We first describe the common experimental setup, followed by the benchmark evaluations and the construction audit.
% We evaluate Science Horizon-lab and SciVLABench through four complementary experiments. First, we compare three representative visuomotor policies on ten certified laboratory tasks using both task-level and step-level metrics. Second, we examine whether learned policies can coordinate robot behavior with externally triggered human interventions. Third, we evaluate object-binding generalization by testing policies on geometrically distinct laboratory objects that are not observed during training. Finally, we conduct a controlled audit of 100 task commands to characterize the first-pass reliability, recoverability, and remaining manual effort of the proposed protocol-to-task compilation framework.

\subsection{Experimental Setup}
\label{sec:experimental-setup}

\subsubsection{Policies and Demonstrations.}
We evaluate three representative visuomotor policies:
$\pi_{0.5}$~\cite{physicalintelligence2025pi05},
Action Chunking with Transformers (ACT)~\cite{zhao2023act},
and Diffusion Policy (DP)~\cite{chi2023diffusionpolicy}.
For each task, we train a separate checkpoint using demonstrations
generated by its certified expert program $\mathcal{P}$.
Unless otherwise specified, each checkpoint is evaluated over
30 independently sampled episodes with randomized initial conditions.

\subsubsection{Policy Inputs and Training.}
Each policy receives four RGB views---three fixed workspace cameras and one wrist-mounted camera---together with robot proprioception.
All policies predict a 7D end-effector command for position displacement, orientation displacement, and gripper control.
Images are resized to $480\times480$ for $\pi_{0.5}$ and $256\times256$ for ACT and DP.
ACT and DP operate at $10$ Hz, with ACT predicting chunks of 100 control steps.

ACT and DP use AdamW for 100K optimization steps. The learning rate is $1\times10^{-4}$, and the batch size is 128.
$\pi_{0.5}$ is fine-tuned for 100K steps with a learning rate of $2.5\times10^{-5}$ and a global batch size of 64.
Privileged simulator states are used only to generate expert demonstrations and evaluate task outcomes; they are not provided as policy inputs.

\subsubsection{Evaluation Metrics.}
We report task success rate (SR), which requires the complete ordered success specification $\mathcal{S}$ to be satisfied:
\begin{equation}
\mathrm{SR}
=
\frac{1}{N_{\mathrm{eval}}}
\sum_{n=1}^{N_{\mathrm{eval}}}
\mathbb{I}
\left[
\operatorname{Verify}
\left(
\mathcal{S},
\xi^{(n)}
\mid
\prec
\right)
=1
\right],
\end{equation}
where $N_{\mathrm{eval}}=30$, $\xi^{(n)}$ is the execution trace of episode $n$, and $\prec$ denotes the required ordering among success conditions.

To measure partial procedural completion, we additionally report
step-wise success rate (SSR):
\begin{equation}
\mathrm{SSR}
=
\frac{1}{N_{\mathrm{eval}}}
\sum_{n=1}^{N_{\mathrm{eval}}}
\frac{1}{m}
\sum_{i=1}^{m}
z_i^{(n)},
\end{equation}
where $z_i^{(n)}\in\{0,1\}$ indicates whether condition $s_i$ is
satisfied in episode $n$ after all of its procedural prerequisites under
$\prec$ have been satisfied, and $m$ is the number of ordered evaluation
steps. SR measures complete task execution, whereas SSR measures the
fraction of the intended procedure completed before failure. Both are shown as percentages.
% The additional metrics and protocols used for human-in-the-loop coordination and held-out object generalization are introduced in their corresponding subsections.

\subsection{Benchmark Evaluation}
\label{sec:benchmark-evaluation}

We evaluate \BenchName\ along three complementary dimensions: policy performance across heterogeneous laboratory procedures, coordination with externally triggered interventions, and generalization to held-out object bindings.
Beyond ranking policies, these experiments examine whether the benchmark can expose procedural, coordination, and transfer failures that are obscured by aggregate task success alone.

\subsubsection{Policy Comparison and Procedural Diagnosis}
\label{sec:main-policy-evaluation}

We evaluate $\pi_{0.5}$, ACT, and Diffusion Policy on ten certified tasks spanning the five operation families defined in Section~\ref{sec:benchmark}.
The tasks range from basic object interaction to precision-critical and multi-stage laboratory manipulation.

\begin{table}[t]
    \centering
    \vspace{2mm}
    \caption{Same-task policy evaluation on ten tasks grouped by operation family. Task names are shortened for space. }
    \label{tab:main-policy-results}
    \small
    \setlength{\tabcolsep}{2.2pt}
    \renewcommand{\arraystretch}{1.1}
    \begin{tabular*}{\columnwidth}{@{\extracolsep{\fill}}lcrrrrrr@{}}
    \toprule
    \multirow[c]{2}{*}[-0.5ex]{Task}
& \multirow[c]{2}{*}[-0.5ex]{Steps}
    & \multicolumn{2}{c}{$\pi_{0.5}$}
    & \multicolumn{2}{c}{ACT}
    & \multicolumn{2}{c}{DP} \\
    \cmidrule(lr){3-4}
    \cmidrule(lr){5-6}
    \cmidrule(lr){7-8}
    & & SR & SSR & SR & SSR & SR & SSR \\
    \midrule

        \rowcolor{gray!15}
        \multicolumn{8}{c}{
        \textit{Apparatus \& Workspace Interaction}} \\
        Place Flask
            & 2 & 86.7 & 88.3 & 83.3 & 85.0 & 56.7 & 60.0 \\
        Open Drawer
            & 2 & 96.7 & 96.7 & 93.3 & 93.3 & 66.7 & 71.7 \\
        Insert Glass Rod
            & 2 & 23.3 & 36.7 & 20.0 & 30.0 & 6.7 & 11.7 \\

        \rowcolor{gray!15}
        \multicolumn{8}{c}{
        \textit{Liquid Handling \& Transfer}} \\
        Pour Solution
            & 2 & 50.0 & 63.3 & 46.7 & 51.7 & 36.7 & 53.3 \\

        \rowcolor{gray!15}
        \multicolumn{8}{c}{
        \textit{Solid Handling \& Weighing}} \\
        Pick Up Tube
            & 2 & 63.3 & 81.7 & 76.7 & 78.3 & 43.3 & 43.3 \\
        Unscrew Bottle
            & 2 & 16.7 & 58.3 & 26.7 & 56.7 & 6.7 & 31.7 \\

        \rowcolor{gray!15}
        \multicolumn{8}{c}{
        \textit{Thermal Control \& Incubation}} \\
        Heat Beaker
            & 3 & 60.0 & 73.3 & 53.3 & 56.7 & 40.0 & 63.3 \\

        \rowcolor{gray!15}
        \multicolumn{8}{c}{
        \textit{Mixing \& Agitation}} \\
        Shake Tube
            & 2 & 36.7 & 48.3 & 70.0 & 73.3 & 40.0 & 53.3 \\
        Stir with Tool
            & 2 & 16.7 & 25.0 & 23.3 & 28.3 & 3.3 & 6.7 \\
        Transport and Mix
            & 4 & 0.0 & 42.5 & 3.3 & 40.8 & 0.0 & 29.2 \\

        \midrule
        Average
            & -- & 45.0 & 61.4 & 49.7 & 59.4 & 30.0 & 42.4 \\
        \bottomrule
    \end{tabular*}

\end{table}

As shown in Table~\ref{tab:main-policy-results}, ACT achieves the highest average SR, whereas $\pi_{0.5}$ obtains the highest average SSR.
This difference indicates that the policy with the best complete-task performance is not necessarily the one that makes the most procedural progress before failure.
Across the five operation families, performance does not exhibit a clear family-level ordering and instead varies more strongly with task-specific manipulation demands.
In particular, precision-critical tasks such as \emph{Unscrew The Pill Bottle} and \emph{Insert Glass Rod Into Beaker} remain difficult for all three policies despite their short procedural.
% horizons.
% Across the five families, performance does not follow a consistent
% family-level difficulty ordering and instead varies strongly with the
% specific manipulation demands of each task. For example,
% \emph{Apparatus and Workspace Interaction} contains both
% \emph{Open Drawer}, on which $\pi_{0.5}$ and ACT exceed 93\% SR, and
% \emph{Insert Glass Rod}, for which the best SR is only 23.3\%.
% Similarly, the two-step \emph{Unscrew Bottle} task remains difficult
% despite its short procedural horizon. These contrasts show that task
% difficulty is determined not only by family membership or step count,
% but also by precision, object affordances, and the required state
% transitions.

The gap between SR and SSR further reveals failures hidden by terminal success alone.
On \emph{Transport and Mix}, complete success reaches at most 3.3\%, while SSR reaches 42.5\%, indicating that policies often complete early operations before failing at a later stage.
By combining family-based coverage with ordered step-level verification, \BenchName\ supports both comparison across laboratory operation types and diagnosis of where execution breaks down within a procedure.

\subsubsection{Human-Agent Coordination}
\label{sec:human-coordination}

% We next evaluate whether learned policies can coordinate robot behavior with an external intervention whose timing is randomized and hidden from the policy.
% We consider two tasks.
% In \emph{Lamp-Triggered Heating}, the robot must remain inactive until an alcohol lamp is ignited and then move the vessel toward the heating region.
% In \emph{Post-Addition Transport}, the robot must wait until material is added to a vessel before transporting it to the target workstation.
% The policy must infer the intervention from RGB observations rather than privileged state information.
We further evaluate whether learned policies can coordinate robot behavior with external interventions whose timing is randomized and hidden from the policy. We consider two representative scenarios. In \emph{Lamp-Triggered Heating}, the robot must remain inactive until an alcohol lamp is ignited and then move the vessel to the heating region. In \emph{Post-Addition Transport}, the robot must wait until material is added to a vessel before transporting it to the target workstation. In both scenarios, the policy must identify the intervention from RGB observations alone without access to privileged state information.

% To separate pre-intervention safety from post-intervention behavior, we report task success rate (SR), premature violation rate (PVR), and post-trigger response latency.
% PVR measures the proportion of episodes in which the robot contacts or substantially moves a restricted object before the intervention:
To distinguish pre-intervention safety from post-intervention behavior, we report task success rate (SR), premature violation rate (PVR), and post-trigger response latency. PVR measures the fraction of episodes in which the robot contacts or substantially moves a restricted object before the required intervention:
\begin{equation}
\mathrm{PVR}
=
\frac{1}{N_{\mathrm{eval}}}
\sum_{n=1}^{N_{\mathrm{eval}}}
v_n,
\quad v_n\in\{0,1\}.
\end{equation}
where $v_n$ indicates a premature interaction in episode $n$.

% \begin{equation}
% \mathrm{PVR}
% =
% \frac{1}{N_{\mathrm{eval}}}
% \sum_{n=1}^{N_{\mathrm{eval}}}
% \mathbb{I}
% \left[
% \text{a premature interaction occurs in episode } n
% \right].
% \end{equation}
Response latency measures the delay between the intervention trigger and the first valid robot response that initiates the subsequent manipulation stage:
\begin{equation}
\mathrm{Latency} = t^*_{\mathrm{response}}-t_{\mathrm{trigger}},
\end{equation}
where $t_{\mathrm{response}}^{*}$ denotes the timestamp of the first valid response. The latency is averaged over episodes where a valid post-trigger response is observed.
% \begin{equation}
% \mathrm{Latency}
% =
% t_{\mathrm{first\ valid\ response}}
% -
% t_{\mathrm{trigger}}.
% \end{equation}

\begin{table}[t]
   \vspace{2mm}
    \centering
    \caption{Human-agent coordination under randomized and hidden intervention timing. Latency is measured in seconds. Each model--task pair is evaluated over 30 episodes.}
    \label{tab:human-coordination}
    \scriptsize
    \setlength{\tabcolsep}{4pt}
    \renewcommand{\arraystretch}{1.0}
    \begin{tabular}{@{}p{0.30\columnwidth}lccc@{}}
        \toprule
        Task & Policy
        & SR $\uparrow$
        & PVR $\downarrow$
        & Latency (s) $\downarrow$ \\
        \midrule
        \multirow{2}{*}{Lamp-Triggered Heating}
            & ACT & 10.0 & 0.0 & 8.1 \\
            & DP  & 3.3  & 0.0 & 6.8 \\
        \midrule
        \multirow{2}{*}{Post-Addition Transport}
            & ACT & 16.7 & 0.0 & 8.3 \\
            & DP  & 13.3 & 0.0 & 7.9 \\
        \bottomrule
    \end{tabular}
\end{table}

% Table~\ref{tab:human-coordination} reports the results under randomized and hidden intervention timing.
% Both policies achieve zero PVR across the evaluated episodes, indicating that they reliably avoid premature manipulation.
% However, final task success remains below 17\% for both tasks, while observed post-trigger responses occur with mean delays of 6.8--8.3 seconds.
% This contrast shows that conservative waiting alone is insufficient for successful human--robot coordination: a policy must also respond to the intervention and complete the subsequent manipulation.
% By separating pre-trigger safety, post-trigger responsiveness, and final task completion, \BenchName\ exposes coordination failures that a single terminal success metric cannot distinguish.
Table~\ref{tab:human-coordination} reports results under randomized and hidden intervention timing. Both policies achieve zero PVR across all evaluated episodes, demonstrating reliable avoidance of premature manipulation. However, final task success remains below 17\% for both tasks, with valid post-trigger responses occurring after mean delays of 6.8--8.3 seconds. This gap reveals that conservative waiting alone is insufficient for effective human--robot coordination: policies must not only defer actions safely but also detect interventions, respond appropriately, and complete the subsequent manipulation. By disentangling pre-trigger safety, post-trigger responsiveness, and final task completion, \BenchName\ reveals distinct coordination failures that cannot be captured by a single terminal success metric.

\subsubsection{Generalization to Held-Out Object Bindings}
\label{sec:object-generalization}

Same-task evaluation varies initial conditions while retaining the same manipulated object.
We further examine whether policies can transfer to a geometrically distinct object binding that supports the same high-level operation.
We consider two settings: lifting transfers from a beaker to a flask, while transport transfers from a cylinder to a Petri dish.
In each setting, the operation and success semantics remain unchanged, whereas object appearance, geometry, grasp affordances, and contact configurations differ.

We quantify the object-binding generalization gap as the performance drop from seen to held-out bindings:
\begin{equation}
\Delta_{\mathrm{bind}}
=
\mathrm{SR}_{\mathrm{seen}}
-
\mathrm{SR}_{\mathrm{held}},
\end{equation}
where a smaller value indicates stronger retention.
Table~\ref{tab:object-generalization} reports the results on the seen and held-out bindings.

\begin{table}[t]
   \vspace{2mm}
    \centering
    \caption{Held-out object-binding generalization. Each policy is trained only on the seen object and evaluated over 30 randomized episodes for each binding. Results are SR (\%); lower
    $\Delta_{\mathrm{bind}}$ is better.}
    \label{tab:object-generalization}
    \scriptsize
    \setlength{\tabcolsep}{7pt}
    \renewcommand{\arraystretch}{0.95}
    \begin{tabular}{@{}lllcrrr@{}}
        \toprule
        Operation
        & Seen
        & Held-out
        & Policy
        & Seen SR
        & Held-out SR
        & $\Delta_{\mathrm{bind}}$ \\
        \midrule

        \multirow{3}{*}{Lift}
        & \multirow{3}{*}{Beaker}
        & \multirow{3}{*}{Flask}
        & $\pi_{0.5}$ & 83.3 & 80.0 & 3.3 \\
        & & & ACT & 80.0 & 56.7 & 23.3 \\
        & & & DP  & 63.3 & 36.7 & 26.7 \\

        \midrule
        \multirow{3}{*}{Transport}
        & \multirow{3}{*}{Cylinder}
        & \multirow{3}{*}{Petri Dish}
        & $\pi_{0.5}$ & 93.3 & 73.3 & 20.0 \\
        & & & ACT & 86.7 & 33.3 & 53.3 \\
        & & & DP  & 83.3 & 10.0 & 73.3 \\

        % \midrule
        % \multirow{3}{*}{Average}
        % & \multirow{3}{*}{--}
        % & \multirow{3}{*}{--}
        % & $\pi_{0.5}$ & 88.3 & 76.7 & 11.7 \\
        % & & & ACT & 83.3 & 45.0 & 38.3 \\
        % & & & DP  & 73.3 & 23.3 & 50.0 \\
        \bottomrule
    \end{tabular}
\end{table}

% All three policies degrade when the manipulated object is replaced, showing that robustness to randomized instances of a training object does not ensure transfer to a new object binding.
% The degradation is consistently larger for transport than for lifting, suggesting that object changes become more consequential when the policy must maintain a stable grasp throughout a longer manipulation trajectory.
All three policies experience performance degradation when the manipulated object is replaced, indicating that robustness to randomized instances of seen objects does not necessarily translate into generalization to unseen object bindings. The degradation is consistently larger for transport than for lifting, suggesting that object variation has a stronger impact when the policy must maintain stable grasping and control over a longer manipulation trajectory.

Among the evaluated policies, $\pi_{0.5}$ retains substantially more of its seen-object performance, achieving an average held-out SR of 76.7\%, compared with 45.0\% for ACT and 23.3\% for DP.
More broadly, the results demonstrate that the held-out binding protocol isolates sensitivity to object geometry and affordances while preserving the underlying operation, complementing the initial-state variations covered by same-task evaluation.

\subsection{Compilation and Certification Audit}
\label{sec:compilation-audit}

We conduct a controlled audit of 100 task commands to measure the first-pass reliability, autonomous recoverability, and remaining manual effort of \ModelName.
All commands use assets and operations covered by the registered libraries, allowing the audit to isolate task construction and certification within the current system scope rather than open-world capability expansion.

As shown in Table~\ref{tab:compilation-audit}, 58\% of generated tasks pass structural, execution, predicate, and visual-semantic checks on the first attempt. For uncertified candidates, the certification workflow leverages structured diagnostics to trigger two automatic recovery mechanisms: simulator rerunning for transient execution failures and visual-state correction for inconsistent initial configurations. These mechanisms recover an additional 23\% of candidates, raising the fully automated certification rate to 81\%.

This result demonstrates that certification in \ModelName is not merely a passive acceptance filter, but an active refinement process. Through an agentic feedback loop, the system identifies recoverable execution and scene-level inconsistencies and automatically improves generated tasks before benchmark admission. In particular, visual-state correction resolves mismatches between generated simulation scenes and textual task specifications, while predicate and visual-semantic verification jointly ensure that corrected tasks preserve the intended procedural outcomes. This autonomous recovery process substantially improves both task-construction reliability and the scalability of benchmark generation.

The remaining 19\% of candidates fall outside the current automated recovery scope, primarily due to task-specific spatial and interaction configurations not covered by the existing repair mechanisms. These cases can be efficiently corrected through lightweight human intervention, requiring only 1--2 minutes on average per task. This result demonstrates that \ModelName\ enables scalable benchmark construction through reliable automated certification, with lightweight human intervention reserved only for rare task-specific cases.

\begin{table}[t]
   \vspace{2mm}
    \centering
    \caption{
    Controlled compilation and certification audit on 100 task candidates. Recovery categories are mutually exclusive. The cumulative rate shows the proportion certified after each recovery stage.
    }
    \label{tab:compilation-audit}
    \small
    \setlength{\tabcolsep}{4.5pt}
    \renewcommand{\arraystretch}{1.1}
    \begin{tabular}{lrr}
        \toprule
        Certification pathway & Count & Cumulative (\%) \\
        \midrule
        First-pass certification & 58/100 & 58 \\
        Automatic recovery via simulator rerun & 15/100 & 73 \\
        Automatic visual-state correction & 8/100 & 81 \\
        Human correction & 19/100 & 100 \\
        \bottomrule
    \end{tabular}
\end{table}

%% file: sections/06-conclusion.tex
\section{Conclusion}
\label{sec:conclusion}

In this work, we introduced \ModelName, an agentic protocol-to-task compiler that formulated embodied task construction as a compilation problem. By integrating semantic grounding, executable task synthesis, and multi-stage simulation-based certification, \ModelName\ transformed natural-language scientific protocols into semantic-preserving, executable, and verifiable embodied tasks at scale. Based on this framework, we developed \BenchName, a certified benchmark with diverse laboratory scenarios that supports human-in-the-loop interaction, reproducible expert demonstration generation, and fine-grained step-level evaluation. Our experiments demonstrated that current embodied policies still face substantial challenges in scientific experimentation, particularly in long-horizon manipulation and human-agent coordination. Beyond providing a benchmark, \ModelName\ establishes a scalable paradigm for converting scientific knowledge into executable embodied experiences, paving the way for the development and evaluation of future scientific embodied agents.

%% file: sections/appendix.tex
\newpage
\appendix

\setcounter{table}{0}
\setcounter{figure}{0}
\renewcommand{\thetable}{S\arabic{table}}
\renewcommand{\thefigure}{S\arabic{figure}}
% \section{Research Methods}

% \begin{figure*}[t]
%   \centering
%   \includesvg[width=\textwidth]{figures/benchmark_1}
%   \caption{Overview of the SciVLABench.}
%   \Description{}
%   \label{fig:method-overview}
% \end{figure*}

% \clearpage
% \onecolumn

\section{Benchmark Provenance and Operation-Family Assignment}
\label{app:benchmark-provenance}

\subsection{Primary-Family Assignment Principles}
\label{app:family-assignment}

% \BenchName contains 51 protocol-grounded source scenarios curated from scientific literature, automated laboratory systems, wet-lab robotic benchmarks, and laboratory-safety procedures. For corpus-level organization, each source scenario is assigned to exactly one primary operation family according to its dominant embodied operation.
\BenchName comprises 51 protocol-grounded source scenarios curated from scientific literature, automated laboratory systems, wet-lab robotic benchmarks, and laboratory-safety procedures. For benchmark organization, each scenario is assigned to one primary operation family based on its dominant embodied~operation.

% The assignment is based on the physical operation that principally
% realizes the scientific objective, rather than on the disciplinary
% context, the type of laboratory container, or all secondary capabilities
% involved in the procedure. For a multi-stage scenario, the primary
% family is determined by the operation that produces the intended
% experimental outcome. Secondary properties such as sensing,
% safety-critical execution, cultureware handling, long-horizon
% composition, and human intervention are retained as orthogonal metadata
% rather than represented as additional primary families.

% The five primary operation families are defined as follows.

The assignment is determined by the physical operation that primarily realizes the scientific objective, rather than by disciplinary context, laboratoryware type, or the secondary capabilities involved in the procedure. For multi-stage scenarios, the primary family is selected according to the operation that directly leads to the intended experimental outcome. Other properties, including sensing requirements, safety-critical execution, cultureware handling, long-horizon composition, and human intervention, are retained as orthogonal metadata instead of being introduced as additional primary families. The five primary operation families are defined as follows:

\begin{itemize}[leftmargin=*]
    \item \textbf{Liquid Handling and Transfer.} This family covers liquid handling operations, including the aspiration, dispensing, pouring, recovery, and transfer of liquid samples or reagents across containers, instruments, and testing media. Sampling procedures involving pipettes, glass rods, pH paper, or colorimetric dishes are included when liquid transfer constitutes the dominant embodied operation.
    \item \textbf{Mixing and Agitation.} This family covers procedures whose primary objective is to homogenize, redistribute, or agitate materials through stirring, shaking, swirling, or repeated pouring. A procedure is assigned to this family when achieving a mixed or homogenized state, rather than merely transferring material, constitutes the intended experimental outcome.
    \item \textbf{Solid Handling and Weighing.} This family covers the access, manipulation, transfer, and quantitative handling of solid samples and their containers. It includes operations such as weighing, powder transfer, material addition or removal for adjustment, and reagent-container opening when access to the contents is required for subsequent solid-sample handling.
    \item \textbf{Thermal Control and Incubation.} This family covers thermal-control operations, including heating, incubation, drying, temperature monitoring, and post-heating handling, whose primary purpose is to establish, maintain, monitor, or safely terminate a thermal condition. These procedures may involve hot plates, magnetic heating devices, water baths, flames, drying ovens, thermometers, or insulating surfaces.
    \item \textbf{Apparatus and Workspace Interaction.} This family covers embodied operations involving the assembly, configuration, opening, closing, placement, storage, recovery, and organization of laboratory equipment, cultureware, tools, and workspace elements. It also includes safety-related recovery procedures when the dominant embodied operation concerns the state of the apparatus or workspace, rather than liquid handling, solid handling, mixing, or thermal processing.
\end{itemize}

Table~\ref{tab:family-composition} summarizes the family-level composition of the benchmark. The complete scenario-to-family mapping is provided in Table~\ref{tab:scenario-family-assignment}.

\begin{table}[htbp]
    \centering
    \caption{
    Family-wise distribution of the 51 protocol-grounded source
    scenarios in \BenchName.
    }
    \label{tab:family-composition}
    \small
    \begin{tabular}{lrr}
        \toprule
        Primary Operation Family
        & Scenarios
        & Proportion (\%) \\
        \midrule
        Liquid Handling and Transfer
        & 15 & 29.4 \\
        Mixing and Agitation
        & 6 & 11.8 \\
        Solid Handling and Weighing
        & 7 & 13.7 \\
        Thermal Control and Incubation
        & 9 & 17.6 \\
        Apparatus and Workspace Interaction
        & 14 & 27.5 \\
        \midrule
        \textbf{Total}
        & \textbf{51}
        & \textbf{100.0} \\
        \bottomrule
    \end{tabular}
\end{table}

\subsection{Scenario-to-Family Mapping}
\label{app:scenario-family-mapping}

For compact presentation, we use the following abbreviations:
\textbf{LHT} denotes Liquid Handling and Transfer;
\textbf{MA} denotes Mixing and Agitation;
\textbf{SHW} denotes Solid Handling and Weighing;
\textbf{TCI} denotes Thermal Control and Incubation; and
\textbf{AWI} denotes Apparatus and Workspace Interaction.

\setlength{\LTleft}{0pt}
\setlength{\LTright}{0pt}
\setlength{\tabcolsep}{4pt}
\renewcommand{\arraystretch}{1.05}
\begin{longtable}{
    >{\centering\arraybackslash}p{0.10\textwidth}
    P{0.65\textwidth}
    >{\centering\arraybackslash}p{0.15\textwidth}
}
\caption{
Primary-family assignments of the 51 protocol-grounded source
scenarios in \ModelName.
}
\label{tab:scenario-family-assignment} \\
\toprule
Scenario ID
& Source Scenario
& Primary Family \\
\midrule
\endfirsthead

\multicolumn{3}{c}{\tablename\ \thetable\ -- continued from previous page} \\
\toprule
Scenario ID
& Source Scenario
& Primary Family \\
\midrule
\endhead

\midrule
\multicolumn{3}{r}{Continued on next page} \\
\endfoot

\bottomrule
\endlastfoot

S01 & Large-Cylinder-to-Beaker Liquid Transfer & LHT \\
S02 & Medium-Volume Quantitative Liquid Addition & LHT \\
S03 & Pipette-to-Tube Dispensing & LHT \\
S04 & Sequential Dispensing Across Multiple Tubes & LHT \\
S05 & Liquid Recovery from an Erlenmeyer Flask to a Graduated Cylinder & LHT \\
S06 & Funnel-Assisted Container Transfer & LHT \\
S07 & Two-Beaker Rinse Transfer & LHT \\

S08 & Glass-Rod Stirring in a Beaker & MA \\
S09 & Placement and Activation of a Magnetic Stirrer & MA \\
S10 & Circular Mixing of an Erlenmeyer Flask & MA \\
S11 & Tube-Rack Retrieval and Shaking & MA \\
S12 & Repeated Pouring for Liquid Mixing & MA \\
S13 & Gentle Swirling for Uniform Petri-Dish Coating & MA \\

S14 & Heating a Small Beaker on a Hot Plate & TCI \\
S15 & Heated Magnetic Stirring of an Erlenmeyer Flask & TCI \\
S16 & Water-Bath Incubation of a Test Tube & TCI \\
S17 & Flame Heating with Gentle Test-Tube Agitation & TCI \\
S18 & Tripod-Based Beaker Heating over an Alcohol Lamp & TCI \\
S19 & Drying a Petri Dish in a Drying Oven & TCI \\

S20 & Taring and Weighing a Petri Dish & SHW \\
S21 & Pouring Solid Reagent from a Bottle into a Weighing Dish & SHW \\
S22 & Sequential Weighing of Multiple Precursors & SHW \\
S23 & Transferring Weighed Powder into a Beaker & SHW \\
S24 & Unscrewing a Pill-Bottle Cap & SHW \\
S25 & Transferring Aspirin Samples from a Reagent Bottle & SHW \\
S26 & Corrective Material Addition or Removal during Weighing & SHW \\

S27 & Assembly of a Gravity-Filtration Setup & AWI \\
S28 & Precision Insertion of a Test Tube into a Rack & AWI \\
S29 & Returning a Pipette to Its Stand & AWI \\
S30 & Opening a Drying Oven and Loading a Petri Dish & AWI \\
S31 & Retrieving and Storing a Reagent Bottle in a Drawer & AWI \\
S32 & Arranging a Tripod and Heat Source & AWI \\

S33 & Pipette Sampling into a Petri Dish & LHT \\
S34 & Opening and Closing a Petri-Dish Lid & AWI \\
S35 & Petri-Dish Spotting and Lid Replacement & LHT \\
S36 & Transferring a Test-Tube Sample to a Petri Dish & LHT \\
S37 & Transporting a Petri Dish to an Observation or Processing Area & AWI \\
S38 & Pipette Spotting Followed by Tool Storage & LHT \\
S39 & Organizing a Test-Tube Array & AWI \\

S40 & Measuring Beaker Temperature with a Thermometer & TCI \\
S41 & Sampling onto pH Paper with a Pipette & LHT \\
S42 & Transferring a Sample to pH Paper with a Glass Rod & LHT \\
S43 & Mounting pH Paper and Dispensing a Sample & LHT \\
S44 & Dispensing a Test-Tube Sample into a Colorimetric Dish & LHT \\
S45 & Positioning and Returning a Contact pH Probe & AWI \\
S46 & Wiping a Chemical Spill from the Workbench & AWI \\
S47 & Righting a Fallen Erlenmeyer Flask & AWI \\
S48 & Moving a Hot Vessel onto an Insulating Mat & TCI \\
S49 & Safely Removing a Test Tube after Flame Heating & TCI \\
S50 & Closing an Equipment Door and Drawer & AWI \\
S51 & Slow Collision-Aware Transport of Glassware & AWI \\
\end{longtable}

Analytical, sensing, cultureware-handling, and safety-critical properties exhibited by some scenarios are preserved as secondary task metadata rather than used to define primary operation families. For example, thermometer-based temperature measurement is assigned to TCI while retaining a sensing attribute; pH-paper sampling is assigned to LHT while retaining an analytical-testing attribute; and spill cleanup or apparatus recovery is assigned to AWI while retaining a safety-critical attribute. Human intervention is similarly modeled as an orthogonal coordination attribute rather than an additional primary operation family.

% The analytical, sensing, cultureware-handling, and safety-critical
% properties represented by some scenarios are preserved as secondary
% task metadata. For example, thermometer-based temperature measurement
% is assigned to TCI while retaining a sensing tag; pH-paper sampling is
% assigned to LHT while retaining an analytical-testing tag; and spill
% cleanup and apparatus recovery are assigned to AWI while retaining a
% safety-critical tag. Human intervention is likewise represented as an
% orthogonal coordination attribute rather than as a primary operation
% family.

\FloatBarrier
\section{Registered Atomic Skill Library}
\label{app:atomic-skill-library}

% SciHorizon-Lab uses a registered library of 30 atomic manipulation
% skills to translate grounded protocol steps into robot-executable
% programs. Each skill defines a reusable simulator-level operation that
% can be parameterized with grounded entities, target poses, offsets, or
% task-specific execution settings. The skill-planning agent may compose
% multiple atomic skills to realize one protocol step, while a
% deterministic validator ensures that every generated program invokes
% only registered skills with complete and valid arguments.

% Table~\ref{tab:atomic-skill-library} lists the complete atomic skill
% library used in the current implementation. The skill names correspond
% to the callable interfaces exposed by the simulator execution API.

\ModelName leverages a registered library of 30 atomic manipulation skills to compile grounded protocol steps into robot-executable programs. Each skill defines a reusable simulator-level operation parameterized by grounded entities, target poses, spatial offsets, or task-specific execution settings. The skill-planning agent can compose multiple atomic skills to realize a single protocol step, while a deterministic validator ensures that each generated program invokes only registered skills with complete and valid arguments.

Table~\ref{tab:atomic-skill-library} lists the complete atomic skill set used in the current implementation. The skill names correspond to callable interfaces exposed by the simulator execution API.

\setlength{\tabcolsep}{5pt}
\renewcommand{\arraystretch}{1.1}
\begin{longtable}{
    P{0.06\textwidth}
    P{0.25\textwidth}
    P{0.61\textwidth}
}
\caption{
Registered atomic manipulation skills used by SciHorizon-Lab.
Skill names are reported using the exact callable identifiers in
the simulator execution API.
}
\label{tab:atomic-skill-library} \\
\toprule
ID
& Skill
& Functional Description \\
\midrule
\endfirsthead

\multicolumn{3}{c}{\tablename\ \thetable\ -- continued from previous page} \\
\toprule
ID
& Skill
& Functional Description \\
\midrule
\endhead

\midrule
\multicolumn{3}{r}{Continued on next page} \\
\endfoot

\bottomrule
\endlastfoot

1
& \texttt{step\_trajectory}
& Executes a general robot trajectory represented as an ordered
sequence of target waypoints. \\

2
& \texttt{moveto\_entity}
& Moves the robot end effector to a task-defined position above a
specified grounded entity. \\

3
& \texttt{moveto}
& Moves the robot end effector to a specified target position. \\

4
& \texttt{pick}
& Grasps and lifts a specified manipulable object. \\

5
& \texttt{place}
& Precisely places a manipulated object inside a target container
or receptacle. \\

6
& \texttt{drop}
& Releases a manipulated object onto the workspace surface. \\

7
& \texttt{open\_door}
& Opens an articulated door. \\

8
& \texttt{close\_door}
& Closes an articulated door. \\

9
& \texttt{open\_drawer}
& Opens an articulated drawer. \\

10
& \texttt{press}
& Performs a pressing action, such as activating a button or
device control. \\

11
& \texttt{pull}
& Applies a pulling motion to a specified object or articulated
component. \\

12
& \texttt{push}
& Applies a pushing motion to a specified object or articulated
component. \\

13
& \texttt{pour}
& Performs a basic pouring motion by tilting a manipulated
container. \\

14
& \texttt{pour\_to\_entity}
& Moves a source container above a specified target entity and
performs a pouring operation. \\

15
& \texttt{lift}
& Raises the robot end effector or a currently manipulated object. \\

16
& \texttt{reset}
& Returns the robot to its predefined default configuration. \\

17
& \texttt{close\_gripper}
& Closes the robot gripper. \\

18
& \texttt{open\_gripper}
& Opens the robot gripper. \\

19
& \texttt{flip}
& Flips or turns over a manipulated object. \\

20
& \texttt{rotate}
& Rotates the robot wrist or the currently manipulated object. \\

21
& \texttt{open\_laptop}
& Opens the articulated display of a laptop computer. \\

22
& \texttt{wait}
& Keeps the robot inactive for a specified number of simulation
steps. \\

23
& \texttt{wait\_for}
& Keeps the robot inactive until a specified trigger time or
task-defined waiting condition is reached. \\

24
& \texttt{shake}
& Applies a repeated shaking motion to a manipulated object. \\

25
& \texttt{move\_offset}
& Moves the robot end effector by a specified positional offset
relative to its current pose. \\

26
& \texttt{insert\_to\_entity}
& Inserts a manipulated object into the opening or designated
insertion region of a target entity. \\

27
& \texttt{stir\_entity\_with\_tool}
& Uses a manipulated stirring tool to agitate the contents of a
specified container. \\

28
& \texttt{unscrew\_cap}
& Unscrews and removes the cap of a specified container. \\

29
& \texttt{aspirate}
& Aspirates a liquid sample using a pipette, dropper, or compatible
liquid-handling instrument. \\

30
& \texttt{dispense}
& Dispenses or spots an aspirated liquid sample from a pipette,
dropper, or compatible liquid-handling instrument. \\
\end{longtable}

%% file: bibliography/reference.bib
@article{steiner2019organic,
  author  = {Sebastian Steiner and Jakob Wolf and Stefan Glatzel and
             Anna Andreou and Jaros{\l}aw M. Granda and Graham Keenan and
             Trevor Hinkley and Gerardo Aragon-Camarasa and
             Philip J. Kitson and Davide Angelone and Leroy Cronin},
  title   = {Organic Synthesis in a Modular Robotic System Driven by a
             Chemical Programming Language},
  journal = {Science},
  year    = {2019},
  volume  = {363},
  number  = {6423},
  pages   = {eaav2211},
  doi     = {10.1126/science.aav2211}
}

@article{epps2020artificial,
  author  = {Robert W. Epps and Michael S. Bowen and Amanda A. Volk and
             Kameel Abdel-Latif and Suyong Han and Kristofer G. Reyes and
             Aram Amassian and Milad Abolhasani},
  title   = {Artificial Chemist: An Autonomous Quantum Dot Synthesis Bot},
  journal = {Advanced Materials},
  year    = {2020},
  volume  = {32},
  number  = {30},
  pages   = {2001626},
  doi     = {10.1002/adma.202001626}
}

@article{ha2023aidriven,
  author  = {Taesin Ha and Dongseon Lee and Youngchun Kwon and
             Min Sik Park and Sangyoon Lee and Jaejun Jang and
             Byungkwon Choi and Hyunjeong Jeon and Jeonghun Kim and
             Hyundo Choi and Hyung-Tae Seo and Wonje Choi and
             Wooram Hong and Young Jin Park and Junwon Jang and
             Joonkee Cho and Bosung Kim and Hyukju Kwon and Gahee Kim and
             Won Seok Oh and Jin Woo Kim and Joonhyuk Choi and
             Minsik Min and Aram Jeon and Yongsik Jung and Eunji Kim and
             Hyosug Lee and Youn-Suk Choi},
  title   = {{AI}-Driven Robotic Chemist for Autonomous Synthesis of
             Organic Molecules},
  journal = {Science Advances},
  year    = {2023},
  volume  = {9},
  number  = {44},
  pages   = {eadj0461},
  doi     = {10.1126/sciadv.adj0461}
}

@article{dai2024autonomous,
  author  = {Tianwei Dai and Sriram Vijayakrishnan and
             Filip T. Szczypi{\'n}ski and Jean-Fran{\c{c}}ois Ayme and
             Ehsan Simaei and Thomas Fellowes and Rob Clowes and
             Lyubomir Kotopanov and Caitlin E. Shields and Zhengxue Zhou and
             John W. Ward and Andrew I. Cooper},
  title   = {Autonomous Mobile Robots for Exploratory Synthetic Chemistry},
  journal = {Nature},
  year    = {2024},
  volume  = {635},
  number  = {8040},
  pages   = {890--897},
  doi     = {10.1038/s41586-024-08173-7}
}

@inproceedings{li2025chemistry3d,
  author    = {Shoujie Li and Yan Huang and Changqing Guo and Tong Wu and
               Jiawei Zhang and Linrui Zhang and Wenbo Ding},
  title     = {{Chemistry3D}: Robotic Interaction Toolkit for Chemistry
               Experiments},
  booktitle = {2025 IEEE International Conference on Robotics and Automation
               (ICRA)},
  year      = {2025},
  pages     = {8064--8071},
  publisher = {IEEE}
}

@inproceedings{li2025labutopia,
  author    = {Rui Li and Zixuan Hu and Wenxi Qu and Jinouwen Zhang and
               Zhenfei Yin and Sha Zhang and Xuantuo Huang and Hanqing Wang and
               Tai Wang and Jiangmiao Pang and Wanli Ouyang and Lei Bai and
               Wangmeng Zuo and Ling-Yu Duan and Dongzhan Zhou and
               Shixiang Tang},
  title     = {{LabUtopia}: High-Fidelity Simulation and Hierarchical
               Benchmark for Scientific Embodied Agents},
  booktitle = {Advances in Neural Information Processing Systems},
  year      = {2025}
}

@article{liu2026pipette,
  author  = {Zhe Liu and Huanbo Jin and Zhaohui Du and Zhe Wang and
             He Xu and Peijia Li and Jiaming Gu and Quan Lu and Qi Wang and
             Bin Ji and Ting Xiao},
  title   = {An Embodied Simulation Platform, Benchmark, and Data-Efficient
             Augmentation Framework for Wet-Lab Robotics},
  eprint  = {2606.12936},
  archivePrefix = {arXiv},
  primaryClass = {cs.RO},
  year    = {2026}
}

@article{james2020rlbench,
  author  = {Stephen James and Zicong Ma and David Rovick Arrojo and
             Andrew J. Davison},
  title   = {{RLBench}: The Robot Learning Benchmark and Learning
             Environment},
  journal = {IEEE Robotics and Automation Letters},
  year    = {2020},
  volume  = {5},
  number  = {2},
  pages   = {3019--3026},
  doi     = {10.1109/LRA.2020.2974707}
}

@inproceedings{liu2023libero,
  author    = {Bo Liu and Yifeng Zhu and Chongkai Gao and Yihao Feng and
               Qiang Liu and Yuke Zhu and Peter Stone},
  title     = {{LIBERO}: Benchmarking Knowledge Transfer for Lifelong
               Robot Learning},
  booktitle = {Advances in Neural Information Processing Systems},
  year      = {2023},
  volume    = {36},
  pages     = {44776--44791}
}

@inproceedings{gu2023maniskill2,
  author    = {Jiayuan Gu and Fanbo Xiang and Xuanlin Li and Zhan Ling and
               Xiqiang Liu and Tongzhou Mu and Yihe Tang and Stone Tao and
               Xinyue Wei and Yunchao Yao and Xiaodi Yuan and Pengwei Xie and
               Zhiao Huang and Rui Chen and Hao Su},
  title     = {{ManiSkill2}: A Unified Benchmark for Generalizable
               Manipulation Skills},
  booktitle = {International Conference on Learning Representations},
  year      = {2023}
}

@article{mees2022calvin,
  author  = {Oier Mees and Lukas Hermann and Erick Rosete-Beas and
             Wolfram Burgard},
  title   = {{CALVIN}: A Benchmark for Language-Conditioned Policy Learning
             for Long-Horizon Robot Manipulation Tasks},
  journal = {IEEE Robotics and Automation Letters},
  year    = {2022},
  volume  = {7},
  number  = {3},
  pages   = {7327--7334},
  doi     = {10.1109/LRA.2022.3180108}
}

@article{zhang2024vlabench,
  author  = {Shiduo Zhang and Zhe Xu and Peiju Liu and Xiaopeng Yu and
             Yuan Li and Qinghui Gao and Zhaoye Fei and Zhangyue Yin and
             Zuxuan Wu and Yu-Gang Jiang and Xipeng Qiu},
  title   = {{VLABench}: A Large-Scale Benchmark for Language-Conditioned
             Robotics Manipulation with Long-Horizon Reasoning Tasks},
  eprint  = {2412.18194},
  archivePrefix = {arXiv},
  primaryClass = {cs.RO},
  year    = {2024}
}

@inproceedings{wang2024robogen,
  author    = {Yufei Wang and Zhou Xian and Feng Chen and
               Tsun-Hsuan Wang and Yian Wang and Katerina Fragkiadaki and
               Zackory Erickson and David Held and Chuang Gan},
  title     = {{RoboGen}: Towards Unleashing Infinite Data for Automated
               Robot Learning via Generative Simulation},
  booktitle = {Proceedings of the 41st International Conference on
               Machine Learning},
  year      = {2024},
  volume    = {235},
  series    = {Proceedings of Machine Learning Research},
  pages     = {51936--51983},
  publisher = {PMLR}
}

@article{chen2025robotwin2,
  author  = {Tianxing Chen and Zanxin Chen and Baijun Chen and
             Zijian Cai and Yibin Liu and Zixuan Li and Qiwei Liang and
             Xianliang Lin and Yiheng Ge and Zhenyu Gu and
             Weiliang Deng and Yubin Guo and Tian Nian and
             Xuanbing Xie and Qiangyu Chen and Kailun Su and
             Tianling Xu and Guodong Liu and Mengkang Hu and
             Huan-ang Gao and Kaixuan Wang and Zhixuan Liang and
             Yusen Qin and Xiaokang Yang and Ping Luo and Yao Mu},
  title   = {{RoboTwin 2.0}: A Scalable Data Generator and Benchmark
             with Strong Domain Randomization for Robust Bimanual
             Robotic Manipulation},
  eprint  = {2506.18088},
  archivePrefix = {arXiv},
  primaryClass = {cs.RO},
  year    = {2025}
}

@article{ren2026labvla,
  author  = {Baochang Ren and Xinjie Liu and Xi Chen and Yanshuo Liu and
             Chenxi Li and Daqi Gao and Zeqin Su and Jintao Xing and
             Zirui Xue and Rui Li and Xiangyu Zhao and Shuofei Qiao and
             Minting Pan and Wangmeng Zuo and Lei Bai and Dongzhan Zhou and
             Ningyu Zhang and Huajun Chen},
  title   = {{LabVLA}: Grounding Vision-Language-Action Models in
             Scientific Laboratories},
  eprint  = {2606.13578},
  archivePrefix = {arXiv},
  primaryClass = {cs.RO},
  year    = {2026}
}

@inproceedings{brohan2023rt1,
  author    = {Anthony Brohan and Noah Brown and Justice Carbajal and
               Yevgen Chebotar and Joseph Dabis and Chelsea Finn and
               Keerthana Gopalakrishnan and Karol Hausman and
               Alexander Herzog and Jasmine Hsu and others},
  title     = {{RT-1}: Robotics Transformer for Real-World Control at Scale},
  booktitle = {Robotics: Science and Systems XIX},
  year      = {2023},
  doi       = {10.15607/RSS.2023.XIX.025}
}

@inproceedings{zitkovich2023rt2,
  author    = {Brianna Zitkovich and Tianhe Yu and Sichun Xu and Peng Xu and
               Ted Xiao and Fei Xia and Jialin Wu and Paul Wohlhart and
               Stefan Welker and Ayzaan Wahid and others},
  title     = {{RT-2}: Vision-Language-Action Models Transfer Web Knowledge
               to Robotic Control},
  booktitle = {Proceedings of the 7th Conference on Robot Learning},
  year      = {2023},
  volume    = {229},
  series    = {Proceedings of Machine Learning Research},
  pages     = {2165--2183},
  publisher = {PMLR}
}

@inproceedings{driess2023palme,
  author    = {Danny Driess and Fei Xia and Mehdi S. M. Sajjadi and
               Corey Lynch and Aakanksha Chowdhery and Brian Ichter and
               Ayzaan Wahid and Jonathan Tompson and Quan Vuong and
               Tianhe Yu and Wenlong Huang and Yevgen Chebotar and
               Pierre Sermanet and Daniel Duckworth and Sergey Levine and
               Vincent Vanhoucke and Karol Hausman and Marc Toussaint and
               Klaus Greff and Andy Zeng and Igor Mordatch and Pete Florence},
  title     = {{PaLM-E}: An Embodied Multimodal Language Model},
  booktitle = {Proceedings of the 40th International Conference on
               Machine Learning},
  year      = {2023},
  volume    = {202},
  series    = {Proceedings of Machine Learning Research},
  pages     = {8469--8488},
  publisher = {PMLR}
}

@article{king2009automation,
  author  = {Ross D. King and Jem Rowland and Stephen G. Oliver and
             Michael Young and Wayne Aubrey and Emma Byrne and
             Maria Liakata and Magdalena Markham and Pinar Pir and
             Larisa N. Soldatova and Andrew Sparkes and Kenneth E. Whelan and
             Amanda Clare},
  title   = {The Automation of Science},
  journal = {Science},
  year    = {2009},
  volume  = {324},
  number  = {5923},
  pages   = {85--89},
  doi     = {10.1126/science.1165620}
}

@article{szymanski2023alab,
  author  = {Nathan J. Szymanski and Bernardus Rendy and Yuxing Fei and
             Rishi E. Kumar and Tanjin He and David Milsted and
             Matthew J. McDermott and Max Gallant and Ekin Dogus Cubuk and
             Amil Merchant and Haegyeom Kim and Anubhav Jain and
             Christopher J. Bartel and Kristin Persson and Yan Zeng and
             others},
  title   = {An Autonomous Laboratory for the Accelerated Synthesis of
             Inorganic Materials},
  journal = {Nature},
  year    = {2023},
  volume  = {624},
  number  = {7990},
  pages   = {86--91},
  doi     = {10.1038/s41586-023-06734-w}
}

@inproceedings{todorov2012mujoco,
  author    = {Emanuel Todorov and Tom Erez and Yuval Tassa},
  title     = {{MuJoCo}: A Physics Engine for Model-Based Control},
  booktitle = {2012 IEEE/RSJ International Conference on Intelligent
               Robots and Systems},
  year      = {2012},
  pages     = {5026--5033},
  publisher = {IEEE},
  doi       = {10.1109/IROS.2012.6386109}
}

@inproceedings{savva2019habitat,
  author    = {Manolis Savva and Abhishek Kadian and Oleksandr Maksymets and
               Yili Zhao and Erik Wijmans and Bhavana Jain and
               Julian Straub and Jia Liu and Vladlen Koltun and
               Jitendra Malik and Devi Parikh and Dhruv Batra},
  title     = {Habitat: A Platform for Embodied {AI} Research},
  booktitle = {Proceedings of the IEEE/CVF International Conference on
               Computer Vision},
  year      = {2019},
  pages     = {9339--9347},
  doi       = {10.1109/ICCV.2019.00943}
}

@inproceedings{shen2021igibson,
  author    = {Bokui Shen and Fei Xia and Chengshu Li and
               Roberto Mart{\'i}n-Mart{\'i}n and Linxi Fan and
               Guanzhi Wang and Claudia P{\'e}rez-D'Arpino and
               Shyamal Buch and Sanjana Srivastava and Lyne P. Tchapmi and
               Micael E. Tchapmi and Kent Vainio and Josiah Wong and
               Li Fei-Fei and Silvio Savarese},
  title     = {{iGibson} 1.0: A Simulation Environment for Interactive
               Tasks in Large Realistic Scenes},
  booktitle = {2021 IEEE/RSJ International Conference on Intelligent
               Robots and Systems},
  year      = {2021},
  pages     = {7520--7527},
  publisher = {IEEE},
  doi       = {10.1109/IROS51168.2021.9636667}
}

@inproceedings{deitke2022procthor,
  author    = {Matt Deitke and Eli VanderBilt and Alvaro Herrasti and
               Luca Weihs and Jordi Salvador and Kiana Ehsani and
               Winson Han and Eric Kolve and Ali Farhadi and
               Aniruddha Kembhavi and Roozbeh Mottaghi},
  title     = {{ProcTHOR}: Large-Scale Embodied {AI} Using Procedural
               Generation},
  booktitle = {Advances in Neural Information Processing Systems},
  year      = {2022},
  volume    = {35},
  pages     = {5982--5994}
}

@article{physicalintelligence2025pi05,
  author  = {{Physical Intelligence} and Kevin Black and Noah Brown and
             James Darpinian and Karan Dhabalia and Danny Driess and
             Adnan Esmail and Michael Equi and Chelsea Finn and
             Niccolo Fusai and Manuel Y. Galliker and Dibya Ghosh and
             Lachy Groom and Karol Hausman and Brian Ichter and
             Szymon Jakubczak and Tim Jones and Liyiming Ke and
             Devin LeBlanc and Sergey Levine and Adrian Li-Bell and
             Mohith Mothukuri and Suraj Nair and Karl Pertsch and
             Allen Z. Ren and Lucy Xiaoyang Shi and Laura Smith and
             Jost Tobias Springenberg and Kyle Stachowicz and
             James Tanner and Quan Vuong and Homer Walke and
             Anna Walling and Haohuan Wang and Lili Yu and Ury Zhilinsky},
  title   = {{$\pi_{0.5}$}: A Vision-Language-Action Model with
             Open-World Generalization},
  eprint  = {2504.16054},
  archivePrefix = {arXiv},
  primaryClass = {cs.RO},
  year    = {2025}
}

@inproceedings{zhao2023act,
  author    = {Tony Z. Zhao and Vikash Kumar and Sergey Levine and
               Chelsea Finn},
  title     = {Learning Fine-Grained Bimanual Manipulation with
               Low-Cost Hardware},
  booktitle = {Proceedings of Robotics: Science and Systems},
  year      = {2023},
  address   = {Daegu, Republic of Korea},
  month     = jul,
  doi       = {10.15607/RSS.2023.XIX.016}
}

@inproceedings{chi2023diffusionpolicy,
  author    = {Cheng Chi and Siyuan Feng and Yilun Du and Zhenjia Xu and
               Eric Cousineau and Benjamin Burchfiel and Shuran Song},
  title     = {Diffusion Policy: Visuomotor Policy Learning via
               Action Diffusion},
  booktitle = {Proceedings of Robotics: Science and Systems},
  year      = {2023},
  address   = {Daegu, Republic of Korea},
  month     = jul,
  doi       = {10.15607/RSS.2023.XIX.026}
}
